\documentclass[]{fairmeta}

\title{WearableQA: A Benchmark for Health Reasoning over Real-World Wearable Data}

\author[1,2,*, \dagger]{Ji Soo Lee}
\author[1,\dagger]{Xilun Chen}
\author[1]{Pierce Chuang}
\author[1]{Ashish Shenoy}
\author[1]{Jason Wei}
\author[1,3,*]{Dohwan Ko}
\author[2]{\protect\\Hyunwoo J. Kim}
\author[1]{Benoit Corda}

\affiliation[1]{Meta}
\affiliation[2]{KAIST}
\affiliation[3]{Korea University}

\contribution[*]{Work done during internship at Meta}
\contribution[\dagger]{Equal contribution}

\usepackage{booktabs}
\usepackage{pifont}
\usepackage{colortbl}
\usepackage{adjustbox}
\usepackage{makecell}
\usepackage{tikz}
\usepackage{wrapfig}
\usepackage{listings}
\usepackage{tcolorbox}
\usepackage{xcolor}

\lstdefinestyle{repr}{
  basicstyle=\ttfamily\scriptsize,
  breaklines=true, columns=fullflexible,
  frame=single, rulecolor=\color{gray!55},
  xleftmargin=1em, framexleftmargin=1em, showstringspaces=false,
  aboveskip=4pt, belowskip=2pt,
}

\newcommand{\cm}{\textcolor{black}{\ding{51}}}
\newcommand{\xm}{\textcolor{black!65}{\ding{55}}}

\abstract{
Recent advances in wearable sensing enable continuous monitoring of physiological and behavioral signals, yet existing benchmarks rarely evaluate whether AI systems can reason over a real user's longitudinal wearable record.
We introduce \benchmarkbb{}, a benchmark comprising 4,084 10-option multiple-choice questions constructed from the wearable time series, blood biomarkers, and demographics of 200 real users, each with up to 500 days of daily measurements.
\benchmark{} preserves authentic wearable distributions that include device noise and inter-individual variability.
To evaluate distinct reasoning capabilities, we introduce 16 question types organized along two complementary axes: data versus health reasoning, which distinguishes computation over longitudinal measurements from physiological interpretation; and single- versus cross-signal reasoning, which separates reasoning about individual signals from the integration of multiple signals.
To construct reliable questions at scale, we adopt a dual-grounding framework that combines literature-grounded physiological findings with statistically validated population-grounded physiological patterns.
This enables the capture of meaningful relationships observed in real-world wearable data.
Evaluation of 14 proprietary and open-source LLMs demonstrates that \benchmark{} effectively differentiates model capabilities, with performance ranging from 19.6\% to 72.9\% against a 10\% chance baseline.
Moreover, \benchmark{} remains far from solved: most models achieve accuracies below 60\%.
Overall, \benchmark{} provides a realistic and diagnostic benchmark for evaluating LLM reasoning over real-world wearable data. 
}

\date{\today}
\correspondence{\email{jislee@kaist.ac.kr}, \email{xilun@meta.com}, \email{ashishvs@meta.com}, \email{corda@meta.com}}

\metadata[Data]{\url{https://github.com/facebookresearch/WearableQA}}
\metadata[Huggingface]{\url{https://huggingface.co/datasets/facebook/WearableQA}}

\begin{document}
\newcommand{\benchmark}{\texttt{WearableQA}}
\newcommand{\benchmarkbb}{\texttt{\textbf{WearableQA}}}
\definecolor{highlight}{HTML}{F0F0F0}
\definecolor{grey}{HTML}{F0F0F0}
\definecolor{highlight-blue}{HTML}{CAE9FF}

\maketitle
\begin{figure}[h!]
    \centering
    \includegraphics[
        width=\linewidth]{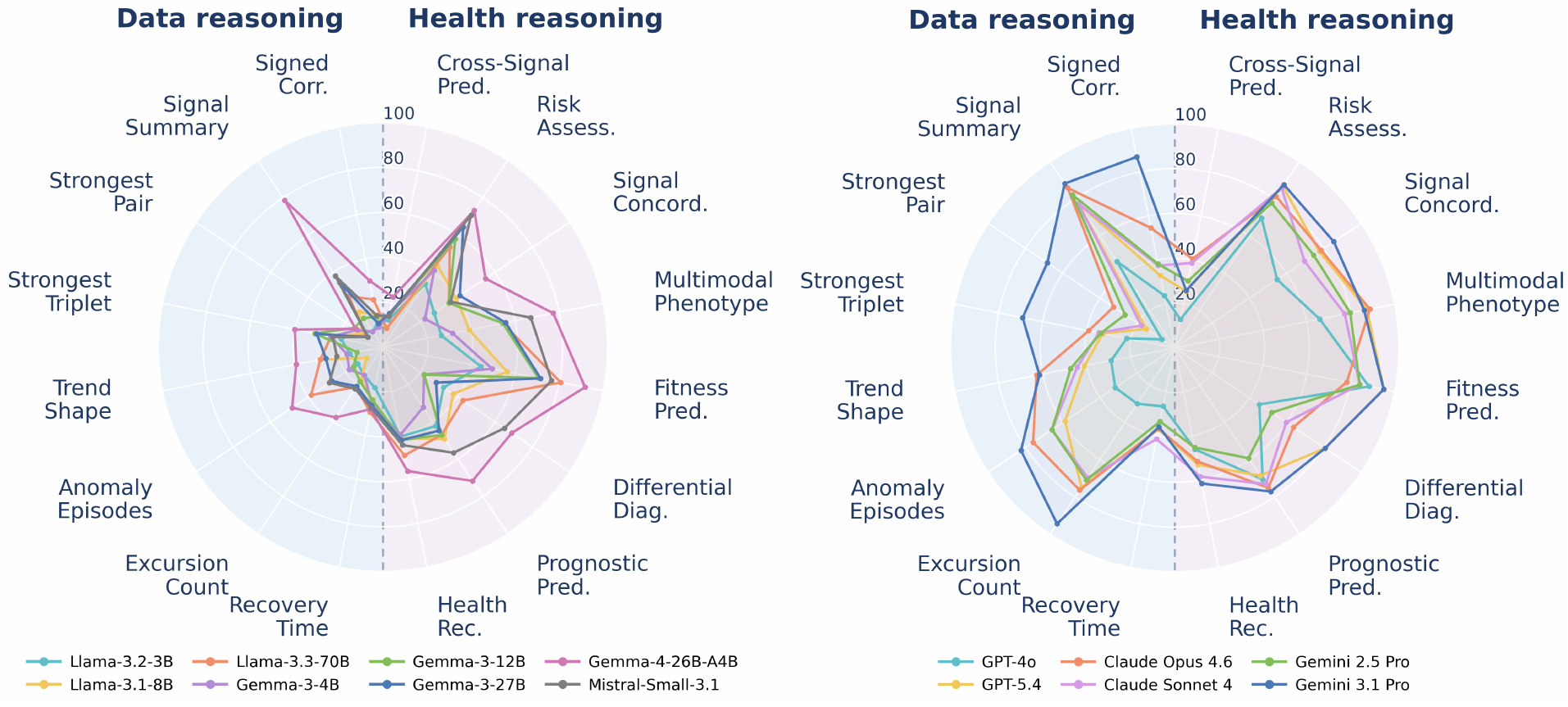}
 \caption{\textbf{Radar plot for performance of Open-Source models (left) and proprietary models (right) on our \benchmark{} Benchmark.}
  }
  \label{fig:web}
\end{figure}

\section{Introduction}
\label{section:intro}
\begin{wrapfigure}{r}{0.45\textwidth}
    \vspace{-\baselineskip}
    \centering
    \includegraphics[width=\linewidth]{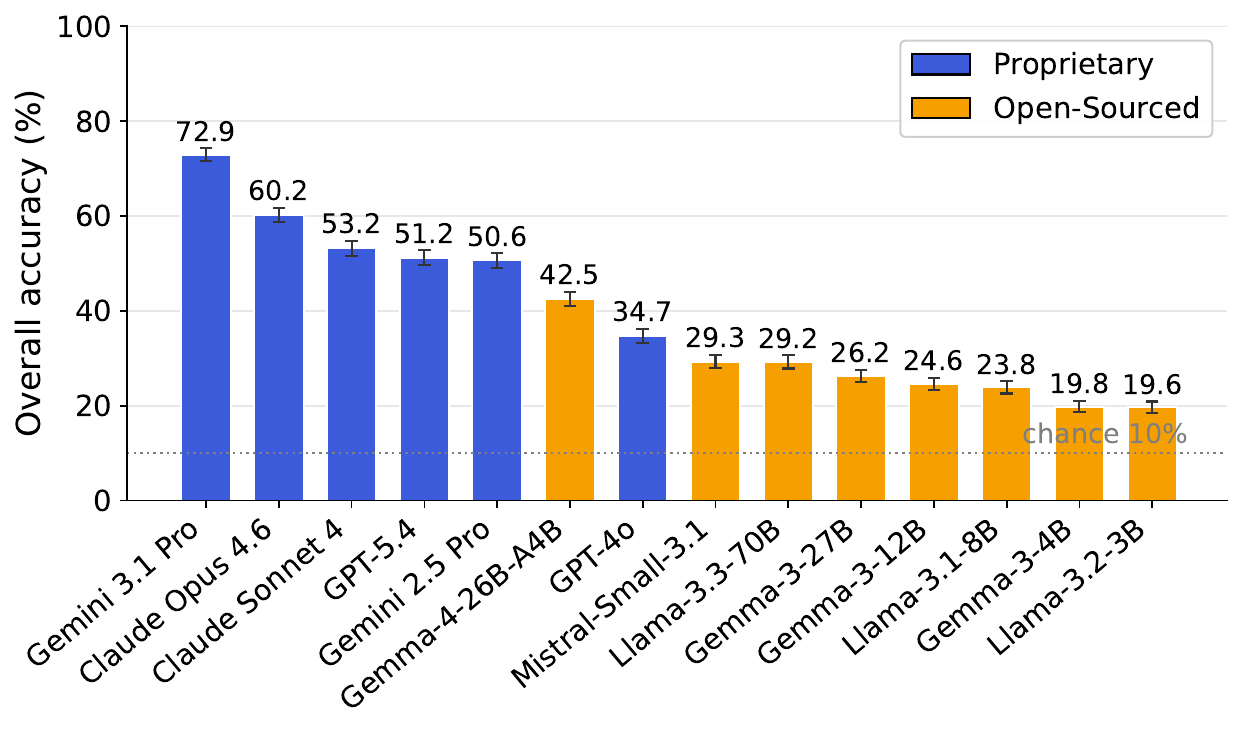}
    \caption{\textbf{Overall accuracy by model (CoT, 95\% CI).}}
    \label{fig:overall}
  \vspace{-0.2cm}
  \end{wrapfigure}

Recent advances in wearable sensing technologies have enabled continuous monitoring of physiological and behavioral signals, including heart rate, sleep, physical activity, and heart rate variability.
Combined with the rapid progress of large language models (LLMs) \citep{hurst2024gpt, singh2025openai, anthropic2025claude4, google2026gemini3, comanici2025gemini, team2024gemma, grattafiori2024llama}, daily longitudinal multimodal measurements are increasingly used to support personalized health understanding, lifestyle assessment, and health monitoring~\citep{kim2024health, merrill2026transforming, khasentino2025personal, pillai2025time2lang}.
Reasoning with wearable data, however, requires more than simply recalling general health knowledge.
The model needs to first identify and compute the information from the noisy longitudinal physiological measurements, and then interpret it in the context of health, which can also be user-specific.
Yet, the ability of LLMs to reason over longitudinal wearable data remains underexplored, with only recent studies beginning to investigate this capability~\citep{merrill2026transforming, jing2026tsaqa, gwiazda2026timeseriesexamagent}.
Systematic benchmarks for evaluating this capability are also scarce.
Existing ones largely rely on synthetic or simulated signals rather than real-world wearable measurements.
Consequently, it remains unclear whether current LLMs can reason about the daily longitudinal wearable history of a \emph{real user} and determine what it may imply about the user's health.
This limitation is increasingly important as LLM-based health assistants continue to emerge \citep{narayanswamy2025scaling, kim2024health, narayanswamy2026towards, merrill2026transforming}.

To this end, we introduce \benchmarkbb, a benchmark for data and health reasoning on real-user longitudinal wearable records. 
\benchmark{} contains 4,084 multiple-choice questions constructed from 200 real users' wearable measurements collected over hundreds of days per individual, which also integrates complementary blood biomarkers and demographic information.
Each instance is grounded in real-user measurements, which naturally capture the inter-individual variability, user-specific baselines across different time spans, and signals. 
\benchmark{} is structured to evaluate model capabilities along two axes: reasoning type and signal complexity.
Reasoning types include data and health reasoning, where \textit{data reasoning} 
computes trends, anomalies, and relationships from the raw wearable measurements, while \emph{health reasoning} requires clinical and physiological interpretation.
Second, the signal complexity axis distinguishes \textit{single-signal} questions from \textit{cross-signal} questions, where the latter challenge the model to integrate patterns and information from two or more signals.
Together, these axes enable a fine-grained diagnosis of whether model failure arises from the computation over longitudinal measurements, health-related interpretation, or the integration of different signals.

Given the nature of the domain, constructing reliable reasoning questions from longitudinal wearable data at scale is challenging.
The questions need to reflect physiological patterns that genuinely arise in real-world data, while allowing the available wearable measurements to be deterministically annotated according to those patterns.
To address these challenges, we adopt a dual-grounding framework composed of two complementary sources: Peer-reviewed literature-grounded findings and statistically validated population-grounded patterns from a large cohort.
Then these objectives are instantiated on each user's measurements through deterministic computations that are refined through careful audit and feedback loops.
Overall, \benchmark{} comprises 16 question types, evenly divided between data reasoning and health reasoning questions.

Our experiments on \benchmark{} across 14 proprietary and open-weight models show that the benchmark effectively distinguishes models with widely varying levels of performance, ranging from 72.9\% to 19.6\% against a 10\% chance baseline.
\benchmark{} also enables a fine-grained diagnosis of model capabilities across reasoning type and signal complexity.
We observe that models generally fall short in data reasoning when it comes to deriving answers directly from the raw measurements.
In addition, cross-signal reasoning remains challenging, with stronger models showing clear drops relative to single-signal reasoning questions, while open-source models generally perform substantially lower on both subsets.

\vspace{1pt}
To sum up, our contributions are summarized as follows:
\begin{itemize}
\item We introduce \benchmark{}, a benchmark of 4,084 multiple-choice questions grounded in the longitudinal wearable records of 200 real users, together with blood biomarkers, demographic attributes, and cohort-relative context.

\item We propose a dual-grounding construction framework that combines literature-grounded physiological findings with population-grounded physiological patterns to derive deterministic ground truth from real user data, together with systematic quality-control and auditing procedures to ensure reliable and non-trivial reasoning questions.

\item We introduce a diagnostic taxonomy of 16 question types spanning two complementary axes: data reasoning versus health reasoning, and single-signal versus cross-signal reasoning.

\item We provide a comprehensive evaluation of proprietary and open-source LLMs, demonstrating that \benchmark{} is discriminative and diagnostic of distinct reasoning limitations over real-world wearable data.

\end{itemize}

\begin{figure}[t!]
    \centering
    \includegraphics[
        width=\linewidth,
    ]{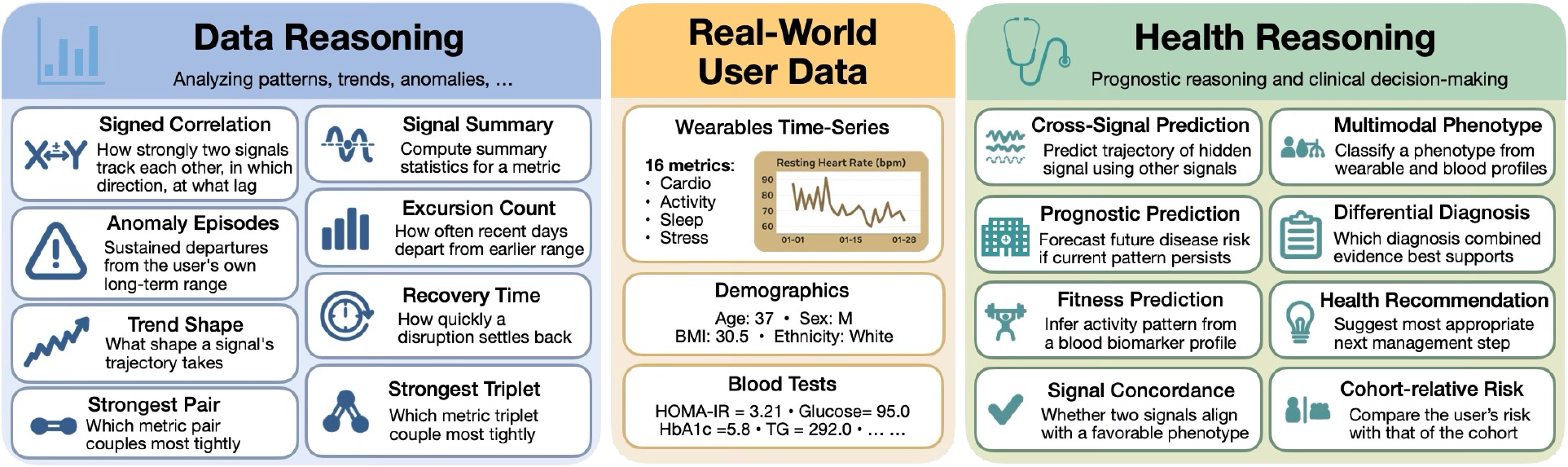}
 \caption{\textbf{Overall taxonomy of \benchmark{}.} 
 Taxonomy for 4,084 questions of real-world user data that span blood, demographics, and time-series wearables.
 The benchmark covers 16 question types of data-reasoning and health-reasoning. 
  }
  \label{fig:taxonomy}
\end{figure}

\section{Related Work}
\subsection{Health-related Benchmarks}
A growing body of work evaluates language models on medical and health reasoning. Benchmarks such as MedQA~\citep{jin2021disease}, MedMCQA~\citep{pal2022medmcqa}, and PubMedQA~\citep{jin2019pubmedqa} primarily assess medical knowledge through examination-style questions and biomedical text, while more recent benchmarks such as HealthBench~\citep{arora2025healthbench} extend evaluation to more realistic health-related interactions. 
Despite their importance, these benchmarks are largely text-based and focus on static questions or cases, providing limited insight into whether models can reason over an individual's longitudinal physiological history.
ECG-QA~\citep{oh2023ecg} moves beyond purely textual inputs by incorporating physiological measurements through ECG question answering.
However, reasoning over a waveform differs from reasoning over longitudinal wearable measurements, where the models must aggregate measurements across time and integrate multiple behavioral and physiological signals; this setting remains unexplored.
More recently, several studies have begun to evaluate LLM reasoning over wearable measurements~\citep{kim2024health, jing2026tsaqa, merrill2026transforming, gwiazda2026timeseriesexamagent}. 
For instance, PHIA~\citep{merrill2026transforming} explores questions over wearable data, but its objective evaluation is based on simulated user trajectories and primarily focuses on retrieval and aggregation over the measurements.
Existing wearable evaluations also typically focus on wearable signals alone, without jointly incorporating complementary health information such as blood biomarkers.
\benchmark{} extends this setting by evaluating reasoning over real-user longitudinal wearable data together with blood biomarkers and demographic information.

\subsection{Time-Series Reasoning Benchmarks}
Recent benchmarks have extended LLM evaluation to time-series reasoning that includes numerical reasoning, forecasting, and question answering over sequential data~\citep{cai2024timeseriesexam, jing2026tsaqa, merrill2026transforming}.
These works provide systematic evaluation beyond static textual inputs, but largely rely on synthetic or simulated time-series signals rather than longitudinal wearable measurements from real users that require interpretation in the context of health.
As a result, they do not capture the complexity of real-world longitudinal wearable measurements or evaluate health interpretation grounded in physiological signals. 
For example, TimeSeriesExam~\citep{cai2024timeseriesexam} evaluates general time-series understanding through procedurally generated synthetic signals spanning pattern recognition, noise understanding, similarity analysis, anomaly detection, and causality.
Consequently, they provide limited evaluation of whether models can both derive information from real-world wearable measurements and interpret those patterns physiologically.
\benchmark{} targets this setting using real longitudinal wearable measurements collected from real users, together with blood biomarkers and demographic information.
It jointly evaluates data reasoning and health reasoning through a diagnostic taxonomy of 16 question types, with deterministic ground truth constructed from literature-grounded physiological findings and statistically validated population-grounded patterns.
\section{The \benchmark \ Benchmark}

\subsection{Real-World User Data}
\begin{figure}[t!]
    \centering
    \includegraphics[
        width=\linewidth,
    ]{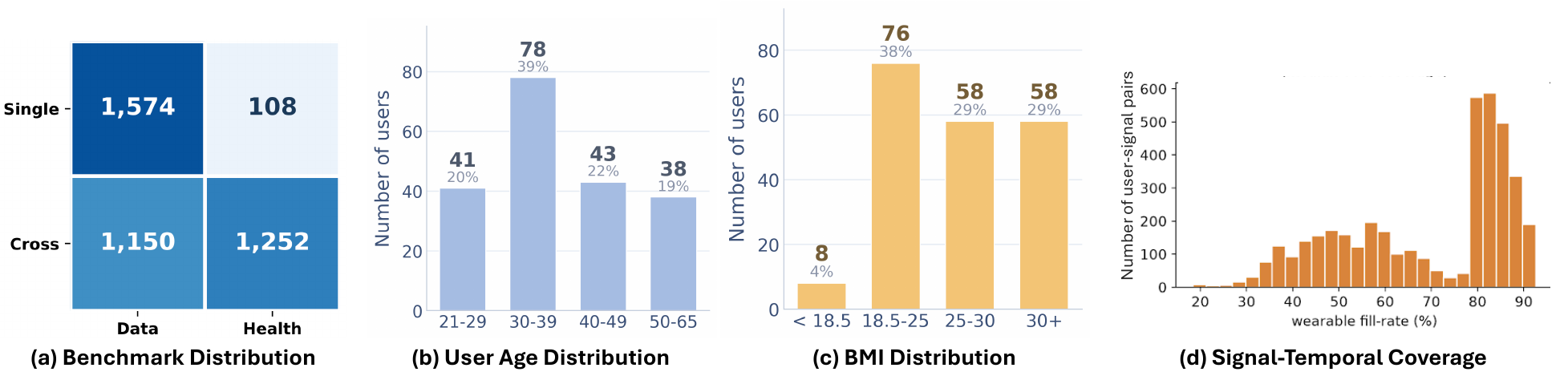}
 \caption{\textbf{Overall statistics of 200 real users from \benchmark{}.}
   }
  \label{fig:user-statistics}
\end{figure}

\benchmark{} comprises longitudinal wearable and blood-panel records from 200 real users, sampled from a larger cohort of in-the-wild wearable time-series data with high coverage across taxonomies.
These 200 users span a diverse range of demographic characteristics (See Fig.~\ref{fig:user-statistics}).

\noindent\textbf{Signals}. Each user contributes a longitudinal multivariate time series comprising 16 daily wearable metrics across four physiological domains. 
Cardio-fitness metrics include resting heart rate, heart-rate variability, and $\mathrm{VO}_2\mathrm{max}$. Activity and energy metrics include steps, active calorie burn, BMR calories, exercise count, exercise duration, average exercise heart rate, and METs. 
Sleep metrics include duration, efficiency, deep-sleep percentage, REM (rapid eye movement) percentage, and bedtime regularity. 
We also include the average stress level. 
Each question is accompanied by up to the most recent 500 days of longitudinal records, represented as daily aggregated measurements.
Each user is also associated with a 17-biomarker blood panel \textit{e.g.}, insulin, HbA1c.
Demographic attributes include age, sex, BMI, and ethnicity.
To support questions that compare an individual with their peers, we provide
cohort-reference percentiles for five core wearable metrics: resting heart rate, steps, heart-rate variability, sleep duration, and sleep efficiency.
The benchmark questions and user trajectories are constructed from the 200-user benchmark cohort, whereas these reference percentiles are computed over a larger cohort of real-world users.

\subsection{Taxonomy}
\label{sec:taxonomy}

\benchmark{} is organized along a primary $2\times2$ taxonomy that crosses \emph{reasoning type} with \emph{signal complexity}, together with a separate, orthogonal \emph{grounding} axis that records how each question's ground truth is sourced.
The 4,084 questions span 16 question types (see Fig.~\ref{fig:taxonomy}), where there exists a total of 2.7k questions of data reasoning and 1.4k questions for health reasoning.
Among them, 1.7k questions ask models to ground a single metric, while the remaining 2.4k require multi-signal grounding. 
Each question is formulated as a 10-option multiple-choice problem, yielding a chance-level accuracy of 10\%, with answer positions uniformly balanced across the full benchmark.

\noindent\textbf{Reasoning: Data vs.\ Health.}
Data-reasoning questions require computing over the raw sensor measurements, such as trends, anomalies, excursions, recovery times, and cross-signal relationships, with the answer determined by the numbers from the wearable records themselves.
Health-reasoning questions instead require clinical and physiological interpretation, including prognostic framing, recommendations, differential reasoning, phenotyping, and two-signal concordance, layering domain knowledge on top of the signals.

\noindent\textbf{Signal Complexity: single vs.\ cross-metric.}
Single-metric questions concern one signal in isolation, whereas cross-metric questions require integrating two or more signals, for example a coupling between physical activity and resting heart rate, or a multi-signal phenotype.

\noindent\textbf{Grounding: literature vs.\ population.}
Orthogonal to the primary axes, each question also carries a label for how its ground truth was derived, either from a literature finding or from a population-grounded pattern (Sec~\ref{sec:construction}).
\begin{figure}[t!]
    \centering
    \includegraphics[width=0.95\linewidth]{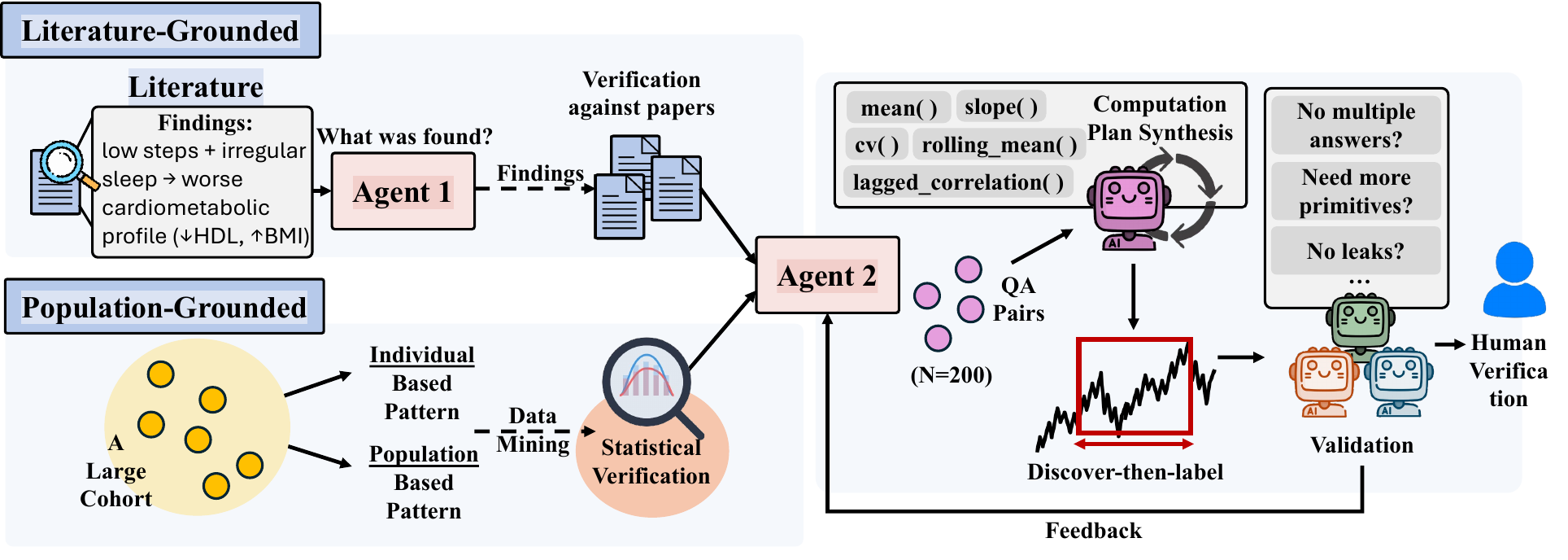}
     \caption{\textbf{Overview of \benchmark{}.} From real users' wearable time series, blood, and
  demographics, we generate questions under dual grounding (peer-reviewed literature findings and statistically verified population-grounded patterns), compute deterministic ground truth from the raw measurements, organize them
  by a $(2\times2)$ reasoning$\times$signal taxonomy, and apply validation loops for quality control.}
  \label{fig:framework}
    \label{fig:overview}
\end{figure}

\subsection{Benchmark Construction}
\label{sec:construction}

Designing questions for wearable health reasoning must consider two objectives.
Questions should (i)~probe clinically meaningful reasoning anchored in biomedical evidence, and (ii)~reflect physiological patterns that genuinely arise in real-world wearable recordings, rather than synthetic scenarios or hand-authored templates.
\benchmark{} reconciles the two through dual grounding: every question is grounded either in the biomedical literature or in a statistically validated pattern discovered directly from a large cohort.
Under both sources, the answer is re-derived from the underlying real measurements rather than asserted a priori, which keeps the benchmark faithful to the target population and resistant to memorization.

\noindent\textbf{Literature-grounded questions.}
Biomedical literature encodes clinically validated relationships among wearable signals, physiological biomarkers, and health outcomes.
We assembled a pool of peer-reviewed consumer-wearable studies and retained only observational studies of those where wearable exposure and clinical outcome are both computable on our cohort, excluding feasibility and modality-mismatched work.
Each retained study was independently retrieved and verified against its primary record \textit{e.g.}, PubMed, and the identifiers were cross-checked across DOI, PMID, and PMCID.
Overall, this yields 11 anchor papers, published in venues such as \emph{Nature Medicine} and \emph{PNAS}, each read at full-text depth rather than from its abstract alone.
To guard against spurious or non-reproducible findings, we further require each retained relationship to be supported by \textbf{at least two independent}, same-direction studies, likewise verified against PubMed/PMC full text.
Each is recorded as a structured card carrying DOI/PMID, verbatim thresholds, cohort details, and an explicit statement of what the source does \emph{not} establish, which prevents downstream questions from over-claiming beyond the evidence.
We deliberately do not import study-specific thresholds or decision boundaries, since such cut-offs are cohort-dependent and rarely transfer across populations.

\noindent\textbf{Population-grounded questions.}
Established literature may not fully cover the diversity of relationships present in large-scale wearable data.
We therefore mine physiological patterns directly from the cohort, surfacing temporal trends, anomalous episodes, recovery dynamics, cross-signal relationships, and higher-order multi-signal interactions, specifically targeting a 28-day window.
A candidate pattern is promoted to a question only if it survives a sequence of predefined statistical-consistency gates evaluated across the cohort, spanning three complementary axes.

\begin{itemize}
\item \textbf{Effect size.} The target must be strong and well separated. 
A coupling must exceed a minimum correlation magnitude ($|\rho| \ge 0.5$) with a consistent sign in both halves of the window, and for comparative questions, the winning option must beat the runner-up by a fixed margin ($\ge 0.15$), so that no answer hinges.

\item \textbf{Robustness.} The labeled outcome must be stable under perturbation. 
It must persist under bootstrap resampling of days (reproducing in $\ge 80\%$ of 30 resamples), under leave-one-out removal of any single day, and, where applicable, across a range of baseline-window choices, so that no single point or knife-edge boundary drives the answer.
Near-boundary cases falling inside an explicit dead-band are rejected.

\item \textbf{Authenticity.} Reproducibility alone does not imply that a pattern is real. 
We therefore apply a cross-user null test, recomputing each statistic on deliberately mismatched user pairings and taking the ratio of null to observed prevalence as an empirical false-discovery rate; a pattern is retained only at $\mathrm{FDR} \le 0.20$, i.e.\ positive predictive value $\ge 80\%$.
\end{itemize}

Importantly, population-grounded questions extend evaluation beyond previously documented clinical findings.
Hence, together, literature-grounded findings and population-grounded patterns provide complementary coverage of wearable reasoning, pairing clinically established knowledge with physiological behavior that emerges directly from real-world data.
\subsection{Discover-then-labeling: Labeling user measurements}
\label{sec:gt}
For labeling the user wearable measurements, we adopt a discover-then-label approach.
Specifically, a reasoning objective is first \textit{discovered} from a literature finding or a population-grounded pattern, and its answer is then \textit{labeled} by running a deterministic program over the individual's real measurements. 

\noindent\textbf{Computation primitives.}
Given a reasoning objective from either a literature finding or a population-grounded pattern, we express the required reasoning as a composition of reusable computation primitives.
For instance, we instantiate a strongest-pair question with \texttt{lagged\_correlation(max\_lag=3)} for each candidate signal pair over lags of up to $\pm3$ days, then conduct \texttt{argmax\_pair()}.
Similarly, a blood-state question applies \texttt{threshold\_flags(glucose, HbA1c, TG, HDL, HOMA\_IR)} to compare each biomarker against predefined clinical cutoffs and aggregates the resulting flags to assign the blood state.
We maintain these operations in a shared primitive library, which allows them to be subsequently reused across question types. 
Representing each instance as an explicit computation graph guarantees that identical inputs always yield identical answers, enabling deterministic and fully reproducible evaluation.

\noindent\textbf{Answer construction.}
The selected primitives are executed on each user's longitudinal wearable trajectory to resolve the answer.
For literature-grounded questions, the computation instantiates the published physiological relationship using personalized baselines and cohort-relative statistics, \textit{e.g.}, resting heart rate elevation will be $\ge1.5$ standard deviations above the user's recent 28-day mean.
A scenario is admitted only when the individual's data deterministically satisfies the underlying criterion, and users for whom it is undefined or ambiguous are discarded rather than approximated.
For population-grounded questions, the answer follows directly from the discovered relationship in the user's own trajectory.
In both cases, the ground truth is computed entirely from the underlying measurements.

\noindent\textbf{Distractor construction.}
Each question presents 10 options; 9 of the incorrect options are constructed differently for the two reasoning types, reflecting what can be verified in each.
Each data-reasoning template defines a distractor category appropriate to the quantity it asks about, such as the opposite trend direction, an adjacent time window, or an incorrect magnitude drawn from the cohort distribution of the same quantity. 
Because the answer is produced by an executable computation, we extend that computation to evaluate all ten options and retain an item only when the gold option uniquely satisfies it. 
Health-reasoning answers involve interpretation rather than a single computed quantity, so their distractors are instead derived from the grounded finding itself: alternative interpretations of the same physiological relationship that are clinically plausible but inconsistent with the individual's measurements. 

\subsection{Benchmark Validation}
\label{sec:validation}

\benchmark\ applies two complementary validation procedures to ensure that instances are correct, reliable, and free of trivial shortcuts.

\noindent\textbf{Computational validity.}
Because every answer is produced by deterministic computation, it can be independently reproduced from the underlying wearable measurements.
Population-grounded questions are retained only after passing predefined statistical consistency gates spanning effect size (a strong, well-separated target), robustness (stability under bootstrap resampling, leave-one-out day removal, and boundary dead-bands), and authenticity (a cross-user null test that bounds the false-discovery rate) so that answers reflect stable physiological relationships rather than spurious observations or single-user artifacts.

\noindent\textbf{Quality control and shortcut auditing.}
Finally, we apply benchmark-level filters: duplicate removal, answer-choice balancing, and distractor verification, and remove any context that would directly reveal the ground truth.
We then use three different models (GPT-5.4, Gemini-3.1-Pro, and Claude-opus-4.6) to review the generated questions, followed by human verification to control the quality of the questions.
As part of the process, we audit for unintended shortcuts.
For instance, in an earlier version of the benchmark, a small number of highly predictable metric pairs, such as active burn with steps, frequently appeared as the strongest relationships.
As a result, models could answer many of the questions even without access to the time series, relying on their prior knowledge.
In other cases, particular answer choices appeared only for certain ground-truth categories, allowing a model to infer the answer from the options alone without examining the wearable data.
Hence, when such issues are identified, we revise the corresponding generation procedure and repeat the verification process.
Together, these steps increase diversity while preventing information leakage and unintended shortcuts.

\begin{figure*}[t]
    \centering
    \includegraphics[width=\linewidth]{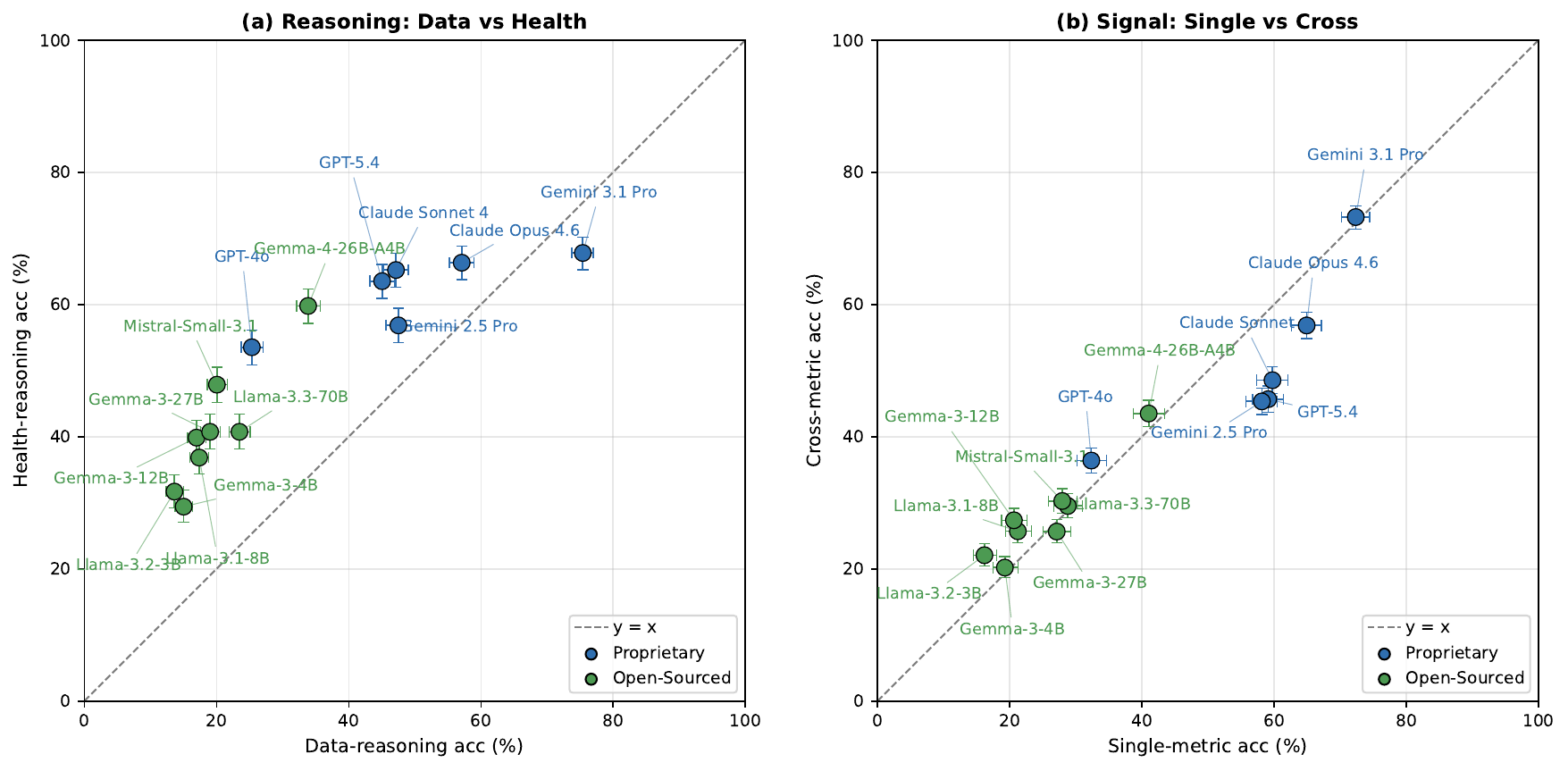}
\caption{\textbf{Accuracy along the two taxonomy axes (CoT).} Each point is a model (95\% CI);
  blue = proprietary, green = open-sourced, dashed line is $y=x$. 
  \textbf{(a)} Most of the models sit above
  $y=x$ (health $>$ data): models reason better about clinical health than about underlying data. 
  \textbf{(b)} Cross-signal questions remain challenging across model scales, while the challenge appears clearer for stronger models (models sit below $y=x$ (single $>$ cross)), while weaker models perform substantially low for both subsets.
}
    \label{fig:intro-2}
\end{figure*}
\section{Experiments}
\label{sec:experiments}
In this section, we first describe our experimental setup and present full results for 14 LLMs, proprietary and open-source, on \benchmark{}, which spans 16 question types.
We then diagnose model capabilities at a finer grain, breaking performance down by reasoning type and signal complexity.
Finally, we present ablation studies, test robustness to input representation, and report an evaluation of agentic models.

\subsection{Experimental Setup}
\label{sec:setup}

We evaluate a diverse set of instruction-tuned LLMs in order to analyze reasoning capability across model families and scales.
Proprietary models include \texttt{GPT-4o} and \texttt{GPT-5.4} (OpenAI), \texttt{Gemini-2.5-Pro} and \texttt{Gemini-3.1-Pro} (Google), and \texttt{Claude-Sonnet-4} and \texttt{Claude-Opus-4.6} (Anthropic).
For open-weight models we evaluate \texttt{Llama-3.2-3B}, \texttt{Llama-3.1-8B}, and \texttt{Llama-3.3-70B} (Meta), \texttt{Gemma-3-4B}, \texttt{Gemma-3-12B}, and \texttt{Gemma-3-27B}, \texttt{Gemma-4-26B-A4B} (Google), and \texttt{Mistral-Small-3.1} (Mistral).
We evaluate every model under the same protocol.
Each question is presented with its associated context, comprising wearable measurements, demographic information, blood biomarkers (when applicable), and cohort reference statistics, following the benchmark definition in Section~\ref{sec:taxonomy}.
Unless otherwise specified, each model receives the user's most recent 500 days of wearable history rendered as a row-wise representation.
By default, the model reasons step by step (chain of thought) before answering, and we additionally report a direct-answer setting in which the model emits the option immediately.
Each response is parsed into a single option identifier and compared against the deterministic ground truth introduced in Section~\ref{sec:gt}.
We report accuracy with 95\% Wilson confidence intervals, and count unparseable or empty responses as incorrect (See Appendix for more details).

\begin{table*}[t] 
\centering        
\caption{\textbf{Comparative Performance of proprietary and open-source models on \benchmark.}
Best and second-best results are shown in \textbf{bold} and \underline{underline}, respectively.}
\label{tab:main}
\footnotesize 
 \begin{adjustbox}{width=0.7\textwidth}
\begin{tabular}{l| cc| cc| c}
\toprule
& \multicolumn{2}{c|}{\textbf{Reasoning}} & \multicolumn{2}{c|}{\textbf{Signal Complexity}} & \\
\cmidrule(lr){2-3} \cmidrule(lr){4-5}
\textbf{Model} & \textbf{Data} & \textbf{Health} & \textbf{Single} & \textbf{Cross} & \textbf{Avg} \\
\hline
\rowcolor{highlight}
\multicolumn{6}{l}{\textit{Proprietary models}} \\
Gemini 3.1 Pro         & \textbf{75.4\,{\scriptsize$\pm$0.8}} & \textbf{67.8\,{\scriptsize$\pm$1.3}} & \textbf{72.4\,{\scriptsize$\pm$1.1}} & \textbf{73.2\,{\scriptsize$\pm$0.9}} & \textbf{72.9\,{\scriptsize$\pm$0.7}}\\
Claude Opus 4.6        & \underline{57.1\,{\scriptsize$\pm$0.9}} & \underline{66.3\,{\scriptsize$\pm$1.3}} & \underline{64.9\,{\scriptsize$\pm$1.2}} & \underline{56.8\,{\scriptsize$\pm$1.0}} & \underline{60.2\,{\scriptsize$\pm$0.8}} \\
Claude Sonnet 4        & 47.1\,{\scriptsize$\pm$1.0} & 65.2\,{\scriptsize$\pm$1.3} & 59.8\,{\scriptsize$\pm$1.2} & 48.5\,{\scriptsize$\pm$1.0} & 53.2\,{\scriptsize$\pm$0.8} \\
GPT-5.4                & 45.0\,{\scriptsize$\pm$1.0} & 63.5\,{\scriptsize$\pm$1.3} & 59.1\,{\scriptsize$\pm$1.2} & 45.7\,{\scriptsize$\pm$1.0} & 51.2\,{\scriptsize$\pm$0.8} \\
Gemini 2.5 Pro         & 47.5\,{\scriptsize$\pm$1.0} & 56.8\,{\scriptsize$\pm$1.3} & 58.1\,{\scriptsize$\pm$1.2} & 45.3\,{\scriptsize$\pm$1.0} & 50.6\,{\scriptsize$\pm$0.8} \\
GPT-4o                 & 25.3\,{\scriptsize$\pm$0.8} & 53.5\,{\scriptsize$\pm$1.4} & 32.3\,{\scriptsize$\pm$1.1} & 36.4\,{\scriptsize$\pm$1.0} & 34.7\,{\scriptsize$\pm$0.7} \\
\hline
\rowcolor{highlight}
\multicolumn{6}{l}{\textit{Open-source models}} \\
Gemma-4-26B-A4B        & \textbf{33.8\,{\scriptsize$\pm$0.9}} & \textbf{59.8\,{\scriptsize$\pm$1.3}} & \textbf{41.0\,{\scriptsize$\pm$1.2}} & \textbf{43.5\,{\scriptsize$\pm$1.0}} & \textbf{42.5\,{\scriptsize$\pm$0.8}} \\
Mistral-Small-3.1      & 20.0\,{\scriptsize$\pm$0.8} & \underline{47.9\,{\scriptsize$\pm$1.4}} & 27.9\,{\scriptsize$\pm$1.1} & \underline{30.3\,{\scriptsize$\pm$0.9}} & \underline{29.3\,{\scriptsize$\pm$0.7}} \\
Llama-3.3-70B          & \underline{23.5\,{\scriptsize$\pm$0.8}} & 40.7\,{\scriptsize$\pm$1.3} & \underline{28.8\,{\scriptsize$\pm$1.1}} & 29.5\,{\scriptsize$\pm$0.9} & 29.2\,{\scriptsize$\pm$0.7} \\
Gemma-3-27B            & 19.0\,{\scriptsize$\pm$0.8} & 40.7\,{\scriptsize$\pm$1.3} & 27.1\,{\scriptsize$\pm$1.1} & 25.6\,{\scriptsize$\pm$0.9} & 26.2\,{\scriptsize$\pm$0.7} \\
Gemma-3-12B            & 17.0\,{\scriptsize$\pm$0.7} & 39.9\,{\scriptsize$\pm$1.3} & 20.6\,{\scriptsize$\pm$1.0} & 27.4\,{\scriptsize$\pm$0.9} & 24.6\,{\scriptsize$\pm$0.7} \\
Llama-3.1-8B           & 17.4\,{\scriptsize$\pm$0.7} & 36.8\,{\scriptsize$\pm$1.3} & 21.2\,{\scriptsize$\pm$1.0} & 25.7\,{\scriptsize$\pm$0.9} & 23.8\,{\scriptsize$\pm$0.7} \\
Gemma-3-4B             & 15.0\,{\scriptsize$\pm$0.7} & 29.4\,{\scriptsize$\pm$1.2} & 19.3\,{\scriptsize$\pm$1.0} & 20.2\,{\scriptsize$\pm$0.8} & 19.8\,{\scriptsize$\pm$0.6} \\
Llama-3.2-3B           & 13.6\,{\scriptsize$\pm$0.7} & 31.7\,{\scriptsize$\pm$1.3} & 16.2\,{\scriptsize$\pm$0.9} & 22.1\,{\scriptsize$\pm$0.8} & 19.6\,{\scriptsize$\pm$0.6} \\
\bottomrule
\end{tabular}
\end{adjustbox}
\end{table*}
\subsection{Main Results}
\label{sec:main-results}

Tab.~\ref{tab:main} and Fig.~\ref{fig:overall} report overall accuracy for all models under the default chain-of-thought protocol.
Accuracy ranges from 19.6\% for \texttt{Llama-3.2-3B}, near the 10\% random baseline, to 72.9\% for \texttt{Gemini-3.1-Pro}, showing substantial variation across model capabilities.
\texttt{Gemini-3.1-Pro} also leads the next-best model, \texttt{Claude-Opus-4.6} (60.2\%), by 12.7 points.
Among open-source models, \texttt{Gemma-4-26B-A4B} performs best at 42.5\%, outperforming \texttt{GPT-4o} by 7.8 points but remains 30.4 points behind the strongest proprietary model.

\noindent\textbf{Data reasoning remains more challenging than health reasoning for most models.}
Decomposing performance by reasoning type reveals a consistent gap between data and health reasoning (Fig.~\ref{fig:intro-2} (left)).
With the exception of \texttt{Gemini-3.1-Pro}, all models perform better on health reasoning.
The gap is particuarly pronounced for \texttt{GPT-4o} with 53.5\% vs. 25.3\%, \texttt{Gemma-4-26B-A4B} with 59.8\% vs. 33.8\%, and \texttt{Mistral-Small-3.1} with 47.9\% vs. 20.0\%.
\texttt{Gemini-3.1-Pro} is the only model to reverse in trend, achieving 75.4\% on data reasoning compared with 67.8\% on health reasoning. 
These results suggest that directly deriving answers from real-user longitudinal measurements remains a major challenge for most models.

\noindent\textbf{Cross-signal reasoning remains challenging across model scales.}
Along the signal-complexity axis, most proprietary models perform better on single-signal than on cross-signal questions (see Fig.~\ref {fig:intro-2} (right)).
The largest gaps are observed for \texttt{GPT-5.4} with 59.1\% vs. 45.7\% and \texttt{Gemini-2.5-Pro} with 58.1\% vs. 45.3\%, followed by \texttt{Claude-Sonnet-4} with 59.8\% vs 48.5\%.
Open-weight models generally perform worse on both subsets, with the strongest model \texttt{Gemma-4-26B-A4B}, reaching $41.0$ on single-signal and $43.5$ on cross-signal questions.
Notably, \texttt{Gemini-3.1-Pro} performs strongly on both subsets, with comparable accuracy on single- and cross-signal questions, that is, 72.4\% vs. 73.2\%.

\noindent\textbf{Per question-type breakdown.}
Fig.~\ref{fig:web} visualizes the accuracy for all 16 question types.
For data reasoning, performance varies substantially across tasks.
\textit{Strongest pair} and \textit{recovery time} are particularly challenging for most models, while \textit{signal summary} and \textit{excursion count} are comparatively easier.
For example, \texttt{Gemini-3.1-Pro} achieves 68.3\% on \textit{strongest pair} and 94.4\% on \textit{excursion count}, but also only 35.9\% on \textit{recovery time}.
Open-source models perform substantially worse on most-data reasoning tasks, with several results close to the 10\% random baseline.
Health-reasoning tasks are generally easier, with strong proprietary models reaching over 80\% on \textit{risk assessment}, \textit{multimodal phenotype}, and \textit{fitness prediction}. 
A notable exception is \textit{cross-signal prediction} for both proprietary and open-source models, where accuracy ranges from 8.3\% for \texttt{Llama-3.1-8B} to only 40.5\% for \texttt{Claude-Opus-4.6}.
Unlike other health-reasoning tasks, this question type requires models to infer one signal from its relationship with another using the user's longitudinal measurements and its knowledge.
Its consistently low performance indicates that grounding cross-signal physiological relationships in user-specific data remains challenging even for the strongest models.

\begin{figure}[t!]
    \centering
    \includegraphics[width=\linewidth]{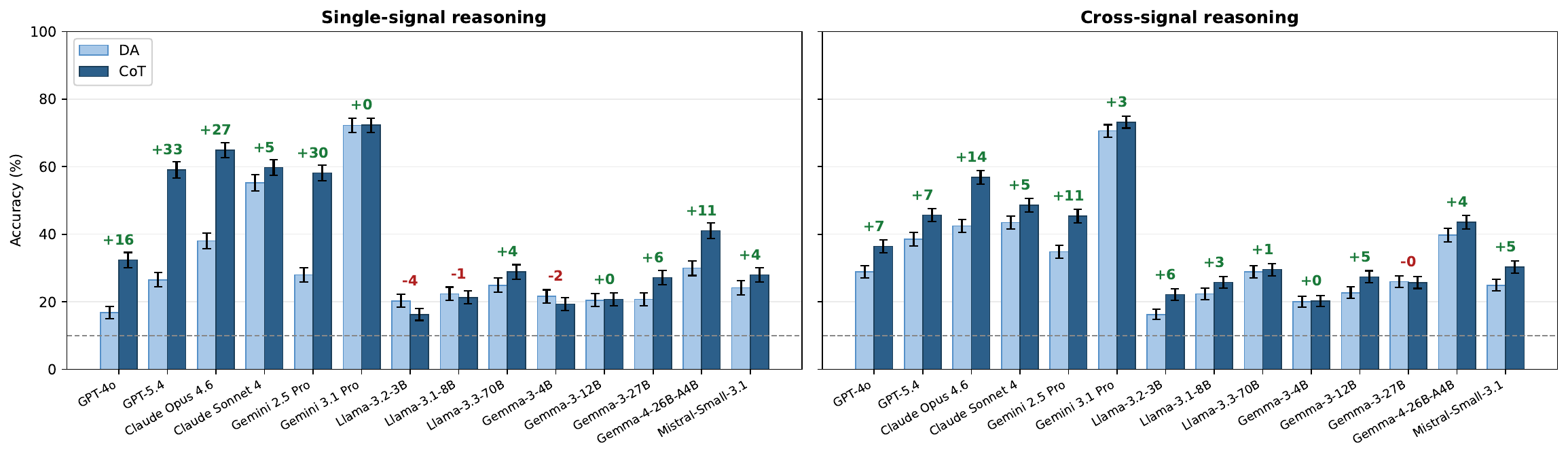}
  \caption{\textbf{Direct-answer vs.\ chain-of-thought across signal complexity.}
  Left is single-signal reasoning, with right cross-signal reasoning, which shows per-model accuracy ($\%$) under direct answering (DA, light) and chain-of-thought (CoT, dark) with 95\% confidence intervals; the number above each pair is $\Delta = \text{CoT} - \text{DA}$ and the dashed line marks the $10\%$ chance level. }
    \label{fig:dacot}
\end{figure}

\noindent\textbf{Chain-of-thought vs.\ direct answer.}
Fig.~\ref{fig:dacot} compares chain-of-thought (CoT) with direct-answer (DA) prompting among the axes of signal complexity.
CoT improves most models, where CoT yields substantially larger improvements for proprietary models, \textit{e.g.}, \texttt{Claude-Opus-4.6} shows gains up to $27$ points for single-signal reasoning and up to $14$ points for cross-signal reasoning.
In contrast, open-weight models show only modest gains under CoT, averaging $+2.3$ points on single-signal and $+3.0$ points on cross-signal.
These results suggest that explicit reasoning substantially improves performance, while its benefit is more limited when reasoning requires integrating information across multiple signals.
See the Appendix for additional results.

\subsection{Analysis}
\label{sec:analysis}

\begin{figure}[t!]
    \centering
    \includegraphics[width=\linewidth]{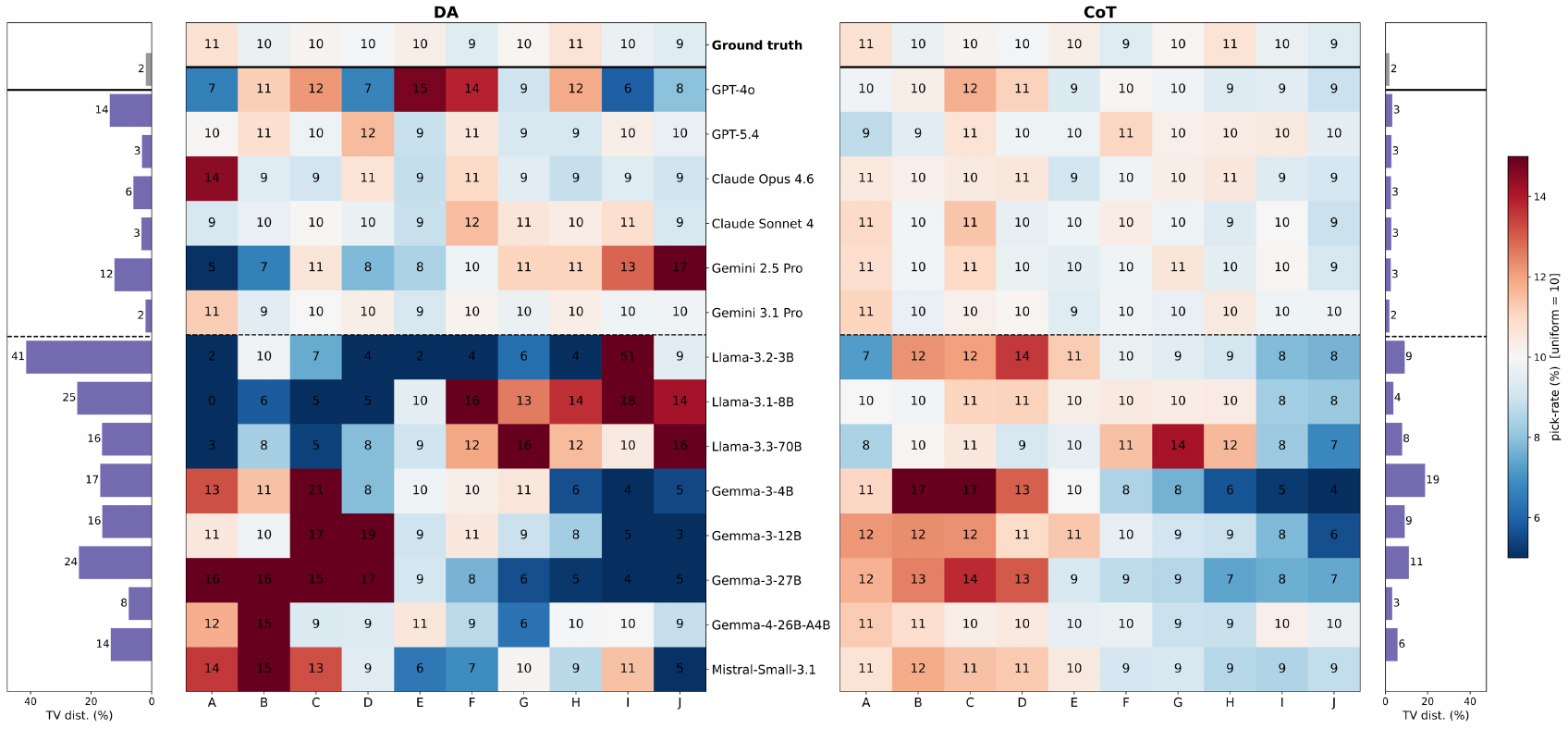}
\caption{\textbf{Answer-letter pick-rate vs.\ uniform ground truth (DA left, CoT right).}
  Cells show each model's pick-rate ($\%$) per letter (red = over-picked, blue = under-picked, white $=10$); side panels give total-variation distance from uniform. 
  Solid rule = ground-truth row, dashed rule = proprietary vs.\ open-sourced. 
  Bias grows from CoT to DA and from larger to smaller models; \texttt{Llama-3.2-3B} under DA collapses onto one letter ($51\%$), while CoT restores near-uniform.}
    \label{fig:bias}
\end{figure}

\begin{table*}[t]
\centering
\begin{minipage}[t]{0.44\textwidth}
\centering
\caption{\textbf{Definitional vs.\ empirical signal pairs.} Definitional pairs are related by construction (same-source signals) and answerable from priors; empirical pairs are physiological couplings that must be read from the data. 
Both groups are restricted to the strong-sync band, ensuring matched relationship strength.
}
\label{tab:ablation-def-vs-emp}
\small
\begin{adjustbox}{width=\linewidth}
\begin{tabular}{l c c c}
\toprule
\textbf{Model} & \textbf{Definitional} & \textbf{Empirical} & \textbf{$\Delta$} \\
\hline
\rowcolor{highlight}
\multicolumn{4}{l}{\textit{Proprietary models}} \\
GPT-4o                 & 50.9\,{\scriptsize$\pm$9.5} & 15.0\,{\scriptsize$\pm$11.4} & +36.0 \\
GPT-5.4                & 63.2\,{\scriptsize$\pm$9.2} & 15.0\,{\scriptsize$\pm$11.4} & +48.2 \\
Claude Opus 4.6        & 70.8\,{\scriptsize$\pm$8.6} & 66.7\,{\scriptsize$\pm$16.0} & +4.0 \\
Claude Sonnet 4        & 80.2\,{\scriptsize$\pm$7.5} & 75.9\,{\scriptsize$\pm$14.3} & +4.4 \\
Gemini 2.5 Pro         & 87.8\,{\scriptsize$\pm$6.1} & 36.3\,{\scriptsize$\pm$16.4} & +51.5 \\
Gemini 3.1 Pro         & 98.2\,{\scriptsize$\pm$1.8} & 94.2\,{\scriptsize$\pm$5.8} & +4.0 \\
\hline
\rowcolor{highlight}
\multicolumn{4}{l}{\textit{Open-source models}} \\
Llama-3.2-3B           & 18.8\,{\scriptsize$\pm$7.4} & 5.8\,{\scriptsize$\pm$5.8} & +13.0 \\
Llama-3.1-8B           & 17.9\,{\scriptsize$\pm$7.2} & 5.8\,{\scriptsize$\pm$5.8} & +12.0 \\
Llama-3.3-70B          & 40.6\,{\scriptsize$\pm$9.3} & 8.9\,{\scriptsize$\pm$8.3} & +31.7 \\
Gemma-3-4B             & 4.6\,{\scriptsize$\pm$3.6} & 8.9\,{\scriptsize$\pm$8.3} & -4.2 \\
Gemma-3-12B            & 17.9\,{\scriptsize$\pm$7.2} & 15.0\,{\scriptsize$\pm$11.4} & +2.9 \\
Gemma-3-27B            & 19.8\,{\scriptsize$\pm$7.5} & 11.9\,{\scriptsize$\pm$10.0} & +7.8 \\
Gemma-4-26B-A4B        & 69.8\,{\scriptsize$\pm$8.7} & 66.7\,{\scriptsize$\pm$16.0} & +3.1 \\
Mistral-Small-3.1      & 24.5\,{\scriptsize$\pm$8.1} & 15.0\,{\scriptsize$\pm$11.4} & +9.5 \\
\bottomrule
\end{tabular}
\end{adjustbox}
\end{minipage}
\hfill
\begin{minipage}[t]{0.54\textwidth}
\centering
\caption{\textbf{Effect of input representation and tool use.}
Accuracy of \texttt{GPT-5.4} under different time-series representations.
Measurements are serialized row-wise, column-wise, as Markdown, or as CSV; image-grid and image-grouped denote separate plots for each signal and range-grouped plots, respectively.
Stats signifies the statistical summary \textit{e.g.}, min, max.
For agentic evaluation, \textit{fixed-repr} provides Python access with the row-wise representation, whereas \textit{multi-view} allows the model to select among the available representations.
}
\label{tab:repr}
\footnotesize
\begin{adjustbox}{width=\linewidth}
\begin{tabular}{l|c|cccc}
\toprule
\textbf{Input condition} & \textbf{Overall} & \textbf{Data} & \textbf{Health} & \textbf{Single} & \textbf{Cross} \\
\midrule
row (baseline)      & 51.2 & 45.0 & 63.5 & 59.1 & 45.7 \\
col                 & 50.3 & 44.0 & 63.1 & 57.9 & 45.0 \\
csv                 & 51.5 & 46.2 & 62.2 & 60.0 & 45.6 \\
markdown            & 49.9 & 43.8 & 62.2 & 57.1 & 44.9 \\
markdown+stats      & 51.9 & 46.7 & 62.4 & 61.1 & 45.5 \\
image-grid          & 36.6 & 25.9 & 58.2 & 32.1 & 39.8 \\
image-grouped       & 34.1 & 24.3 & 53.9 & 29.8 & 37.2 \\
csv+chart           & 51.0 & 45.3 & 62.2 & 58.5 & 45.6 \\
\midrule
\rowcolor{highlight}
\textit{Agentic Evaluation} & & & & & \\
agentic, fixed-repr & 71.3 & 72.9 & 68.1 & 63.8 & 76.5 \\
agentic, multi-view & 69.7 & 71.9 & 65.3 & 60.6 & 76.1 \\
\bottomrule
\end{tabular}
\end{adjustbox}
\end{minipage}
\end{table*}

\noindent\textbf{Positional bias.}
Ground-truth answers of \benchmark{} are balanced across the 10 option positions; an unbiased model should select each about $10\%$ of the time.
As shown in Fig.~\ref{fig:bias}, proprietary models are relatively uniform, whereas smaller models exhibit strong positional preferences, especially under direct answering.
The most extreme case is \texttt{Llama-3.2-3B}, which selects a single option for $51.5\%$ of its direct-answer responses.
Milder preferences also appear in proprietary models: \texttt{Claude-Opus-4.6} favors earlier options, particularly `A', whereas \texttt{Gemini-2.5-Pro} favors later ones like `J'.
Chain-of-thought substantially reduces these biases, \textit{e.g.}, \texttt{Llama-3.2-3B} reduces its maximum option-selection frequency from $51.5\%$ to $13.5\%$, while \texttt{Claude-Opus-4.6} reduces it from $14.5\%$ to $10.6\%$.
This suggests that explicit reasoning helps models overcome positional answer-selection shortcuts, reinforcing that successful performance on \benchmark{} depends on reasoning over the provided evidence rather than exploiting superficial option preferences.

\noindent\textbf{Reliance on priors.}
{To determine whether models rely on prior knowledge when interpreting wearables data, we analyze questions that ask models to assess the strength of association between two signals.
We compare definitional signal pairs, such as step count and active energy expenditure, whose relationship can largely be anticipated with prior knowledge, with empirical pairs, such as resting heart rate and stress level, whose association must be inferred from the observed measurements (Tab.~\ref{tab:ablation-def-vs-emp}).
We restrict both groups to the same strong-coupling band, controlling for relationship strength. 
While some models show only small differences, several models perform substantially better on definitional pairs, including \texttt{Gemini-2.5-Pro} ($+51.5$ points), \texttt{GPT-5.4} ($+48.2$), and \texttt{GPT-4o} ($+36.0$).
This suggests that many models can recognize familiar signal relationships more reliably than they can recover comparably strong, user-specific associations directly from longitudinal measurements.

\noindent\textbf{Effect of input representation and tool use.}
Tab.~\ref{tab:repr} compares time-series representations and Python tool access evaluation where textual formats cover row-wise, column-wise, Markdown, and CSV (see appendix for examples).
Changing the input format provides only marginal gains over the baseline, where the best plain text format of CSV reaches $51.5$ with a $+1.2\%$ gain in data reasoning.
Providing additional summary statistics offers somewhat larger, but limited gains, \textit{e.g.}, markdown$+$stats improves overall accuracy to $51.9$, including a $1.6$ gain in data reasoning and $2.0$ in single-signal reasoning. 
Representing in visuals does not improve performance; the accuracy drops to $36.6$ with separate plots for each signal (image-grid) and further to $34.1$ when signals are grouped by range (image-grouped).
Providing both modalities does not alleviate this limitation: csv$+$chart reaches $51.0$, slightly below csv alone. 
In contrast, providing explicit computational access through an agentic approach substantially improves performance. 
While retaining row-wise representation, enabling the Python access (agentic, fixed-repr) improves by $20.1$ from $51.2$ to $71.3$ overall.
Yet, allowing the agent to additionally select among the available textual and visual representations (agentic, multi-view) provides no further benefit, reaching $69.7$ overall.

\begin{wraptable}{r}{0.46\textwidth}
\vspace{-12pt}
\centering
\caption{\textbf{Ablation of time-series inputs.}
Results are reported for \texttt{Claude-Opus-4.6}. 
Demographics, blood panels, and cohort references are retained in all settings. 
}
\label{tab:ablation-wo-data}
\vspace{-4pt}
\scriptsize
\setlength{\tabcolsep}{3.2pt}
\renewcommand{\arraystretch}{1.08}
\resizebox{\linewidth}{!}{%
\begin{tabular}{ccccc}
\toprule
\textbf{Time series}
& \textbf{History}
& \textbf{Hist.-dep.}
& \textbf{Window-dep.}
& \textbf{Acc.} \\
\midrule
\cm & \cm
& 75.9\,{\scriptsize$\pm$5.7}
& 58.1\,{\scriptsize$\pm$2.1}
& 60.2\,{\scriptsize$\pm$1.5} \\

\cm & \xm
& 39.8\,{\scriptsize$\pm$6.5}
& 58.0\,{\scriptsize$\pm$2.1}
& 60.2\,{\scriptsize$\pm$1.5} \\

\xm & \xm
& 42.6\,{\scriptsize$\pm$6.6}
& 13.7\,{\scriptsize$\pm$1.5}
& 17.3\,{\scriptsize$\pm$1.2} \\

\bottomrule
\end{tabular}%
}
\end{wraptable}
\noindent\textbf{Effect of time-series and history context.}
Removing only the longitudinal history reduces history-dependent question types (\textit{e.g.}, anomaly for baseline calculation) accuracy from $75.9\%$ to $39.8\%$, while leaving window-scoped accuracy nearly unchanged ($58.1\%$ to $58.0\%$).
When all wearable time series are withheld, window-scoped accuracy falls to $13.7\%$, and overall accuracy falls from $60.2\%$ to $17.3\%$.
This selective degradation confirms that \benchmark{} requires models to ground their answers in the relevant wearable measurements.

\section{Conclusion}
\label{sec:conclusion}
We introduced \benchmark{}, a benchmark of 4,084 multiple-choice questions grounded in the longitudinal wearable records, blood biomarkers, and demographics of 200 real users. 
By preserving the noise and inter-individual variability of real-world wearable data, \benchmark{} provides a realistic setting for evaluating whether LLMs can reason over personal health trajectories.
Covering 16 question types, \benchmark{} evaluates models along two complementary axes: data vs. health reasoning and single- vs. cross-signal reasoning, enabling fine-grained diagnosis of the model.
To construct reliable questions at scale, we introduce a dual-grounding framework that combines literature-grounded physiological findings with population-grounded patterns and instantiates them on individual user trajectories.
{Across 14 proprietary and open-source models, performance ranges widely, from 72.9\% accuracy to 19.6\% against a 10\% chance baseline.
Most models generally find data reasoning challenging, having to derive answers directly from the underlying noisy measurements, while cross-signal reasoning remains challenging across model scales.
Overall, we hope \benchmark{} provides a realistic and diagnostic testbed for advancing LLMs that can reason reliably over longitudinal real-world user health data.

\clearpage
\newpage
\bibliographystyle{assets/plainnat}
\bibliography{main}

\newpage
\appendix
\section{Appendix}
\subsection{Evaluation Protocol and Prompt}
\label{sec:eval-protocol}

Every item is scored by \emph{exact match}: the model is shown one user's demographics, wearable time series, blood panel, and a cohort reference distribution, and must select one of ten options (A--J). 
The predicted letter is parsed from the response and compared to the deterministic ground-truth key. All models are queried at temperature $0$.

\noindent\textbf{Prompt structure.}
Each prompt is assembled from a fixed system instruction followed by five context blocks (\textsc{user profile}, \textsc{sensor data}, \textsc{blood biomarker panel}, \textsc{cohort reference}, \textsc{question}/\textsc{options}), and finally an answer-format directive that differs between the chain-of-thought(CoT) and direct-answer conditions.

\noindent\textbf{System instruction (identical in both conditions).}
\begin{quote}\ttfamily\small
You are given a user's demographics, wearable health sensor history, blood biomarker panel, a cohort reference distribution, and a multiple-choice question. 
Analyze the data carefully and select the single best answer.
\end{quote}

\noindent\textbf{Answer-format directive.}
The two evaluation conditions differ \emph{only} in the trailing directive appended after the options; the context blocks are byte-identical, so any accuracy difference is attributable to reasoning elicitation alone.

\begin{table}[h]
\centering
\small
\caption{The only prompt difference between conditions is the closing directive.}
\label{tab:cot-directive}
\begin{tabular}{@{}p{0.20\linewidth}p{0.72\linewidth}@{}}
\toprule
Condition & Trailing directive (verbatim) \\
\midrule
\textbf{With CoT} &
\ttfamily Think step by step.\newline
Show your reasoning, then on the LAST line write ONLY:\newline
Answer: X\newline
where X is the single letter of your answer. \\
\addlinespace
\textbf{Without CoT} &
\ttfamily Choose the single best answer. Respond with ONLY the letter (A-J).
Do not explain or output anything else.\newline
Answer: \\
\bottomrule
\end{tabular}
\end{table}

\noindent\textbf{Answer parsing.}
The response is parsed for \texttt{Answer:\ X} (case-insensitive, optional parentheses); failing that, a final lone letter line, then a leading letter, are accepted as fallbacks. 
Under the no-CoT (direct-answer) the model emits the bare letter directly; under CoT the parser reads the letter off the final line after the reasoning trace.

\subsection{Experimental Details}
\label{supp:details}

\noindent\textbf{Definitional vs Empirical signal pairs.}
To evaluate whether the models rely on their prior knowledge, as shown in Tab~\ref{tab:ablation-def-vs-emp} in the main text, we carefully selected two groups with the same strength-band relationship between the pairs that cover definitional (well-known from prior knowledge, such as step count and active energy expenditure) and empirical (associations referred to from the observed measurements like heart rate and stress level). 
Hence, the final number of samples for each group is 102 and 29 for each definitional and empirical pair, respectively.

\subsection{Time-Series Representations}
\label{sec:representations}

The same wearable history can be handed to the model in several textual serializations or as a rendered plot. 
We evaluate four text formats (\texttt{row}, \texttt{col}, \texttt{markdown}, \texttt{csv}) produced by a single renderer, plus an image condition. We additionally evaluate two hybrid conditions: \texttt{markdown+stats}, which appends summary statistics to the table, and \texttt{csv+chart}, which pairs the best-performing text format with a rendered plot.
All formats carry identical numeric content and identical column labels (\texttt{Steps}, \texttt{RHR}, \texttt{Sleep(h)}, \texttt{HRV}, etc); they differ only in \emph{layout}, which lets us isolate the effect of representation from the effect of information. 
The examples below show a 3-day, 4-signal excerpt.

\noindent\textbf{(a) \texttt{row} --- one line per day (default).}
Each day is a record; signals are inline \texttt{key=value} pairs. 
Emphasizes same-day cross-signal reading.
\begin{lstlisting}[style=repr]
=== SENSOR DATA (row: one line per day) ===
  2023-04-03: Steps=6120.0, RHR=61.0, Sleep(h)=6.80, HRV=42.0
  2023-04-04: Steps=7010.0, RHR=60.0, Sleep(h)=7.10, HRV=45.0
  2023-04-05: Steps=5480.0, RHR=63.0, Sleep(h)=6.40, HRV=39.0
\end{lstlisting}

\noindent\textbf{(b) \texttt{col} --- one block per metric.}
Each signal is a contiguous dated series. Emphasizes within-signal temporal trend/trajectory reading.
\begin{lstlisting}[style=repr]
=== SENSOR DATA (column: one block per metric) ===
  Steps: 2023-04-03=6120.0, 2023-04-04=7010.0, 2023-04-05=5480.0
  RHR: 2023-04-03=61.0, 2023-04-04=60.0, 2023-04-05=63.0
  Sleep(h): 2023-04-03=6.80, 2023-04-04=7.10, 2023-04-05=6.40
  HRV: 2023-04-03=42.0, 2023-04-04=45.0, 2023-04-05=39.0
\end{lstlisting}

\noindent\textbf{(c) \texttt{markdown} --- pipe table.}
Days $\times$ signals grid with a header rule.
\begin{lstlisting}[style=repr]
=== SENSOR DATA (markdown table) ===
| date | Steps | RHR | Sleep(h) | HRV |
|---|---|---|---|---|
| 2023-04-03 | 6120.0 | 61.0 | 6.80 | 42.0 |
| 2023-04-04 | 7010.0 | 60.0 | 7.10 | 45.0 |
| 2023-04-05 | 5480.0 | 63.0 | 6.40 | 39.0 |
\end{lstlisting}

\noindent\textbf{(d) \texttt{csv} --- comma-separated grid.}
The same grid as a header row followed by comma-separated value rows.
\begin{lstlisting}[style=repr]
=== SENSOR DATA (csv) ===
date,Steps,RHR,Sleep(h),HRV
2023-04-03,6120.0,61.0,6.80,42.0
2023-04-04,7010.0,60.0,7.10,45.0
2023-04-05,5480.0,63.0,6.40,39.0
\end{lstlisting}

\noindent\textbf{(e) \texttt{image-grid} --- separate plots for each signal (5 x N).}
\begin{figure*}[h]
    \centering
    \includegraphics[width=0.98\textwidth]{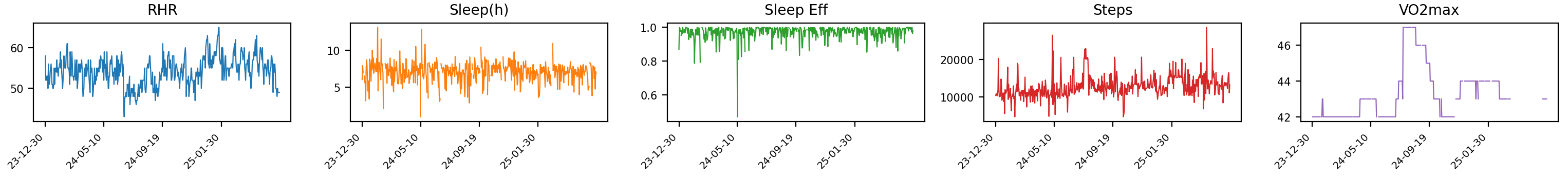}
    \label{suppfig:image-grid}
\end{figure*}

\noindent\textbf{(f) \texttt{image-grouped} --- range-grouped plots.}
\begin{figure*}[h]
    \centering
    \includegraphics[width=0.98\textwidth]{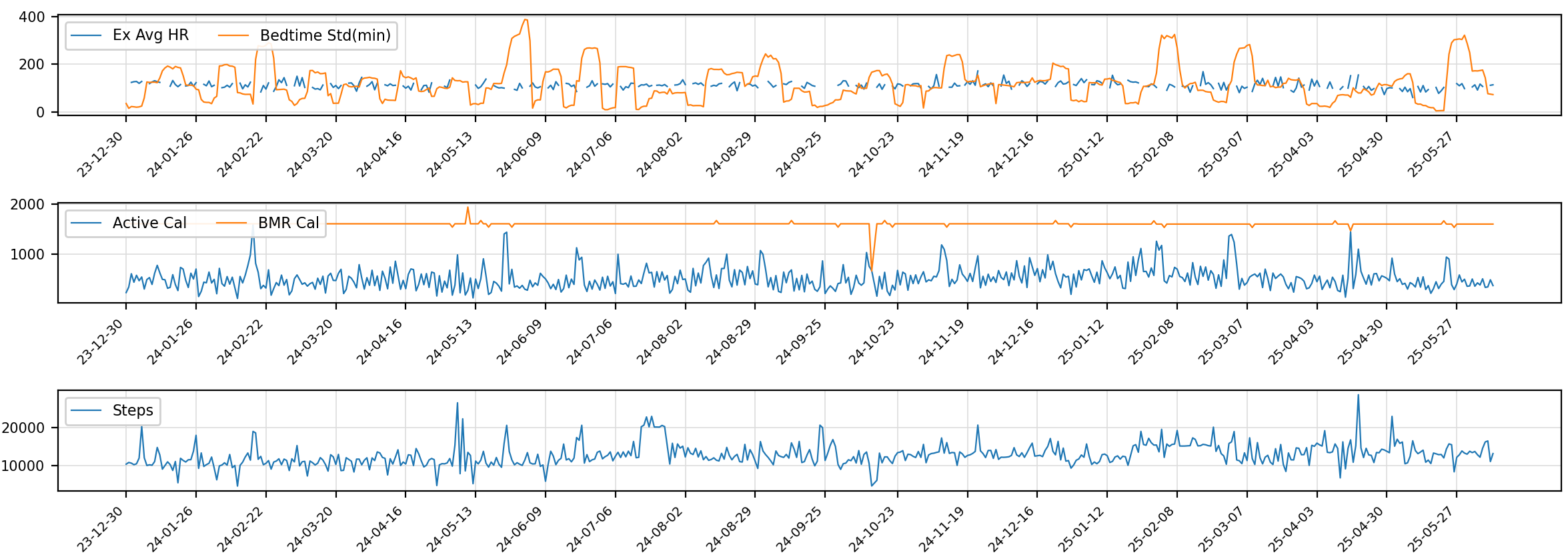}
    \label{suppfig:image-grouped}
\end{figure*}
\clearpage
\begin{figure*}[p]
\centering

\begin{tcolorbox}[
  colback=gray!4,
  colframe=gray!55,
  left=2mm,
  right=2mm,
  boxrule=0.4pt,
  fonttitle=\bfseries,
  title={Example of evaluation prompt (row format, with CoT)}
]
\ttfamily\scriptsize

You are given a user's demographics, wearable health sensor history, blood biomarker panel, a cohort reference distribution, and a multiple-choice question.
Analyze the data carefully and select the single best answer.\\[\baselineskip]

=== USER PROFILE ===\\
\ \ Age: 41, Sex: F, BMI: 24.8, Ethnicity: White\\[\baselineskip]

=== SENSOR DATA (row: one line per day) ===\\
\ \ 2023-04-03: Steps=6120.0, RHR=61.0, Sleep(h)=6.80, HRV=42.0, \dots\\
\ \ 2023-04-04: Steps=7010.0, RHR=60.0, Sleep(h)=7.10, HRV=45.0, \dots\\[\baselineskip]

=== BLOOD BIOMARKER PANEL ===\\
\ \ hdl: 58.0, ldl: 121.0, tg: 96.0, glu: 91.0, hba1c: 5.30, crp: 1.20\\[\baselineskip]

=== COHORT REFERENCE ===\\
\ \ steps: p25=5400, p50=7600, p75=10200\\[\baselineskip]

=== QUESTION ===\\
Based on the most recent 28 days of data, how should this user's step volume be
framed in a metabolic-syndrome health summary?\\[\baselineskip]

=== OPTIONS ===\\
A. \dots\quad B. \dots\quad \dots\quad J. \dots\\[3pt]

Think step by step.\\
Show your reasoning, then on the LAST line write ONLY:\\
Answer: X\\
where X is the single letter of your answer.

\end{tcolorbox}

\caption{The assembled evaluation prompt. Under the no-CoT (Direct Answer) condition, the final instruction is replaced by ``\texttt{Choose the single best answer. Respond with ONLY the letter (A--J). Do not explain or output anything else. Answer:}''.}
\label{fig:eval-prompt}

\end{figure*}
\clearpage
\newpage 
\begin{figure*}[p]
    \centering
    \includegraphics[width=0.98\textwidth]{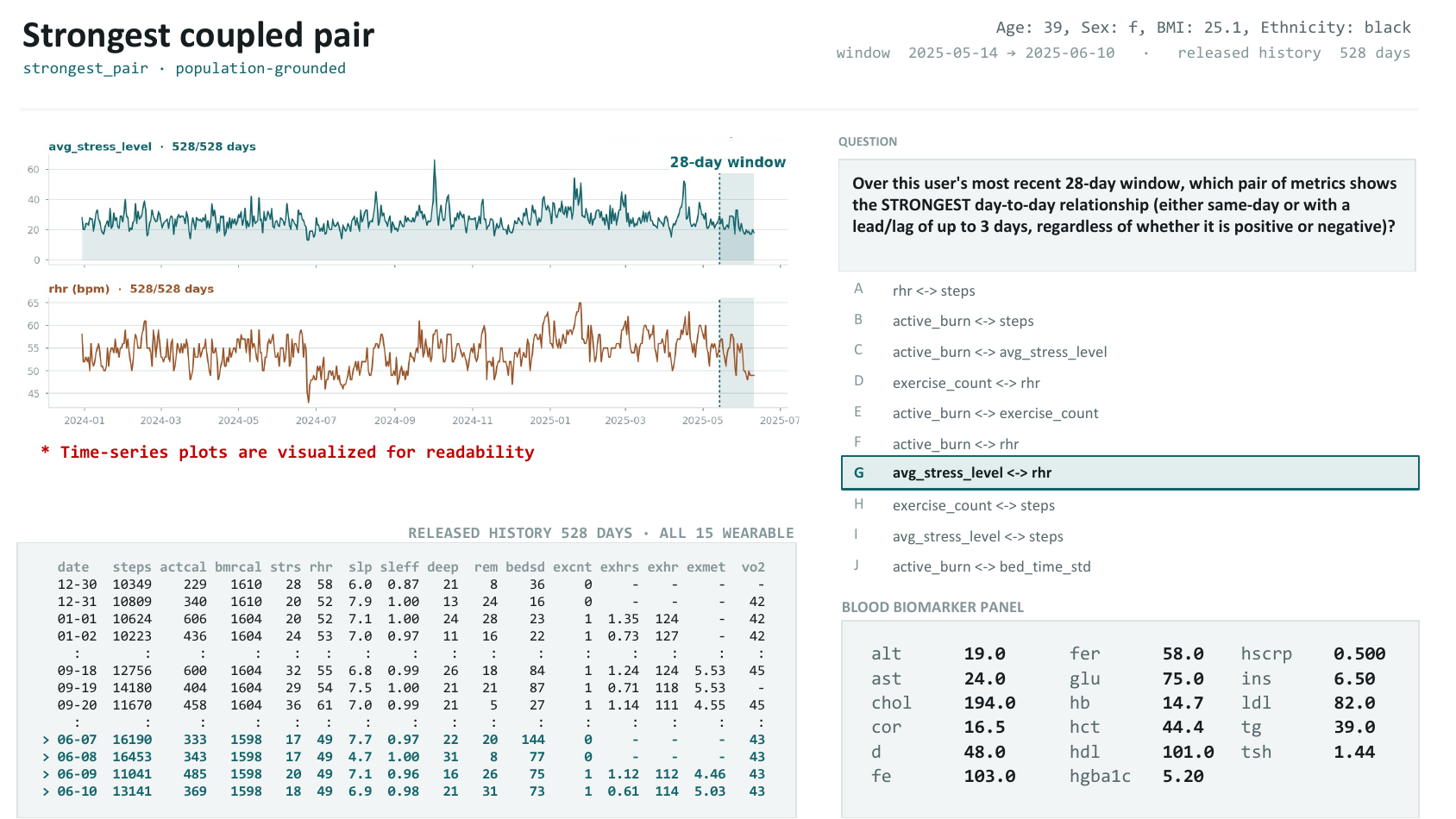}
    \caption{
    \textbf{Example question from \benchmark{} for data reasoning.}
    For readability, the wearable trajectory is visualized as a plot in this figure; models receive the underlying measurements in serialized textual form.
    }
    \label{suppfig:sample1}
\end{figure*}
\begin{figure*}[p]
    \centering
    \includegraphics[width=\textwidth]{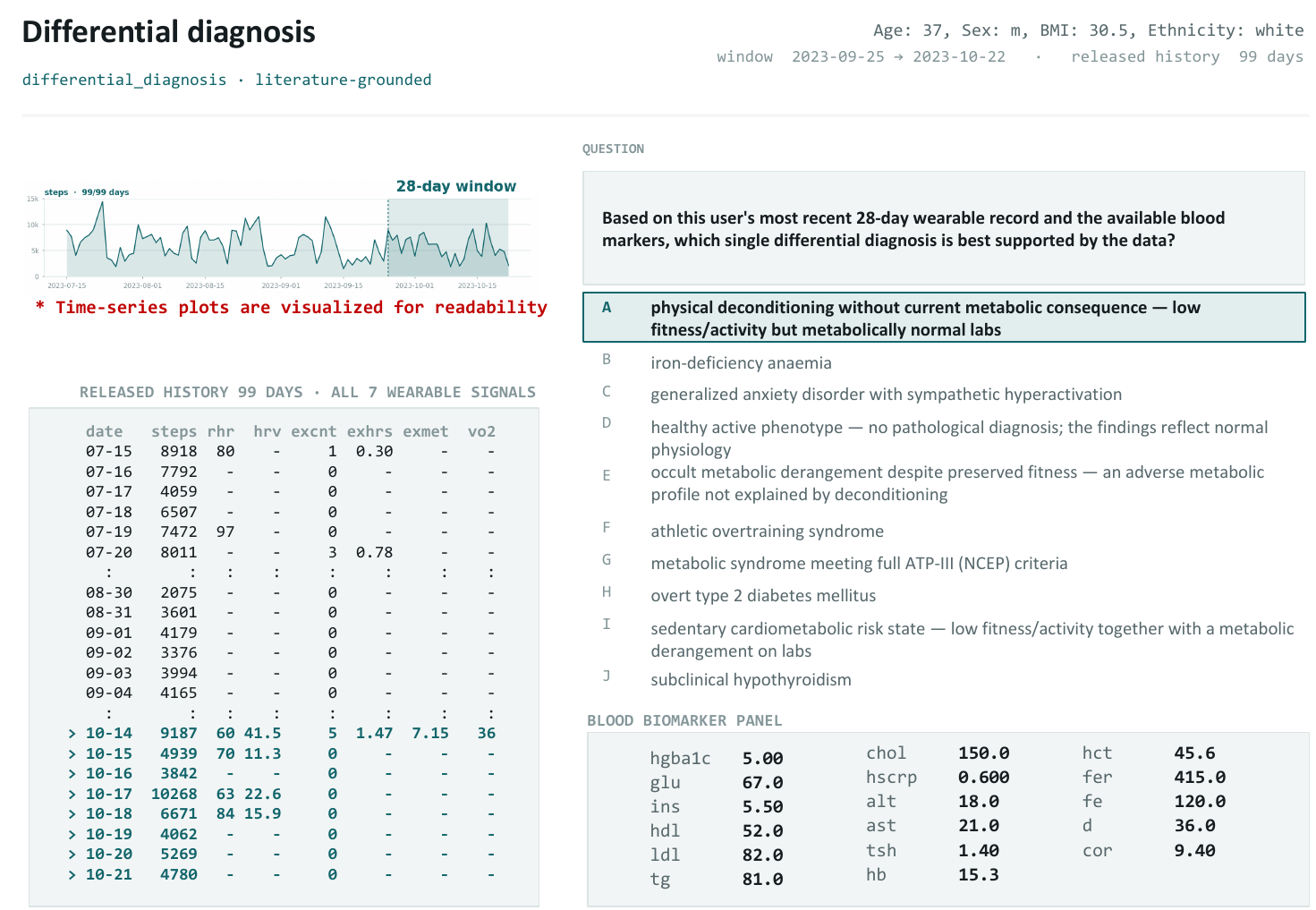}
    \caption{
    \textbf{Example question from \benchmark{} for health reasoning.}
    For readability, the wearable trajectory is visualized as a plot in this figure; models receive the underlying measurements in serialized textual form.
    }
    \label{suppfig:sample2}
\end{figure*}
\begin{table*}[p]
\centering
\caption{\textbf{Per-question-type performance on \benchmark{}.}
Accuracy (\%) across the 16 question types under the default chain-of-thought prompting protocol.
Best and second-best results for each question type are shown in \textbf{bold} and \underline{underline}, respectively.}
\label{tab:per-type}
\footnotesize

\resizebox{\textwidth}{!}{%
\begin{tabular}{l|cccccccc}
\toprule
\multicolumn{9}{c}{\textbf{Data Reasoning}} \\
\midrule
\textbf{Model}
& \textbf{Signed Corr.}
& \textbf{Signal Summary}
& \textbf{Strongest Pair}
& \textbf{Strongest Triplet}
& \textbf{Trend Shape}
& \textbf{Anomaly Episodes}
& \textbf{Excursion Count}
& \textbf{Recovery Time} \\
\midrule
\rowcolor{highlight}
\multicolumn{9}{l}{\textit{Proprietary Models}} \\
GPT-4o
& 23.8$\pm$3.0
& 46.4$\pm$10.7
& 6.9$\pm$2.9
& 21.8$\pm$7.7
& 28.9$\pm$5.7
& 31.9$\pm$6.2
& 30.0$\pm$3.7
& 26.6$\pm$4.1 \\

GPT-5.4
& 32.9$\pm$3.3
& 77.4$\pm$8.9
& 15.3$\pm$4.2
& 32.7$\pm$8.8
& 41.0$\pm$6.2
& 58.8$\pm$6.6
& 74.8$\pm$3.5
& \underline{37.0$\pm$4.5} \\

Claude Opus 4.6
& \underline{54.6$\pm$3.5}
& \underline{85.8$\pm$7.3}
& \underline{32.8$\pm$5.6}
& 39.1$\pm$9.1
& \textbf{62.8$\pm$6.1}
& \underline{75.9$\pm$5.7}
& \underline{76.3$\pm$3.4}
& 36.8$\pm$4.5 \\

Claude Sonnet 4
& 37.5$\pm$3.4
& 77.4$\pm$8.9
& 17.9$\pm$4.5
& 34.5$\pm$8.9
& 44.3$\pm$6.3
& 65.8$\pm$6.3
& 69.7$\pm$3.7
& \textbf{41.5$\pm$4.5} \\

Gemini 2.5 Pro
& 38.2$\pm$3.4
& 82.2$\pm$8.0
& 26.6$\pm$5.2
& 33.6$\pm$8.8
& 47.3$\pm$6.3
& 65.8$\pm$6.3
& 70.9$\pm$3.6
& 33.5$\pm$4.4 \\

Gemini 3.1 Pro
& \textbf{86.9$\pm$2.4}
& \textbf{88.2$\pm$6.7}
& \textbf{68.3$\pm$5.5}
& \textbf{69.1$\pm$8.6}
& \underline{61.5$\pm$6.2}
& \textbf{82.4$\pm$5.0}
& \textbf{94.4$\pm$1.8}
& 35.9$\pm$4.4 \\

\hline
\rowcolor{highlight}
\multicolumn{9}{l}{\textit{Open-source Models}} \\

Llama-3.2-3B
& 11.3$\pm$2.2
& 8.3$\pm$5.6
& 11.7$\pm$3.8
& 19.0$\pm$7.3
& 15.0$\pm$4.5
& 13.4$\pm$4.5
& 15.1$\pm$2.8
& 18.4$\pm$3.6 \\

Llama-3.1-8B
& 14.5$\pm$2.5
& 19.0$\pm$8.3
& 10.9$\pm$3.7
& 23.6$\pm$7.9
& 28.4$\pm$5.7
& 8.8$\pm$3.7
& 17.4$\pm$3.0
& 24.8$\pm$4.0 \\

Llama-3.3-70B
& 21.6$\pm$2.9
& 27.3$\pm$9.5
& 9.1$\pm$3.4
& 23.6$\pm$7.9
& 28.4$\pm$5.7
& 38.4$\pm$6.5
& 21.4$\pm$3.3
& 29.5$\pm$4.2 \\

Gemma-3-4B
& 9.1$\pm$2.0
& 8.3$\pm$5.6
& 13.8$\pm$4.1
& 23.6$\pm$7.9
& 16.3$\pm$4.7
& 18.0$\pm$5.1
& 14.9$\pm$2.8
& 25.5$\pm$4.0 \\

Gemma-3-12B
& 13.9$\pm$2.4
& 15.4$\pm$7.6
& 15.3$\pm$4.2
& 30.9$\pm$8.6
& 11.7$\pm$4.0
& 15.7$\pm$4.8
& 18.4$\pm$3.1
& 23.9$\pm$3.9 \\

Gemma-3-27B
& 10.7$\pm$2.2
& 34.5$\pm$10.1
& 8.0$\pm$3.2
& 30.0$\pm$8.5
& 25.9$\pm$5.5
& 27.3$\pm$5.9
& 21.0$\pm$3.2
& 26.4$\pm$4.1 \\

Gemma-4-26B-A4B
& 30.1$\pm$3.2
& 78.6$\pm$8.7
& 14.6$\pm$4.1
& \underline{40.0$\pm$9.2}
& 39.3$\pm$6.2
& 48.6$\pm$6.7
& 37.7$\pm$3.9
& 28.2$\pm$4.1 \\

Mistral-Small-3.1
& 14.4$\pm$2.5
& 38.1$\pm$10.4
& 8.0$\pm$3.2
& 22.7$\pm$7.8
& 20.9$\pm$5.1
& 28.7$\pm$6.0
& 22.3$\pm$3.3
& 27.5$\pm$4.1 \\

\bottomrule
\end{tabular}
}

\vspace{6pt}

\resizebox{\textwidth}{!}{%
\begin{tabular}{l|cccccccc}
\toprule
\multicolumn{9}{c}{\textbf{Health Reasoning}} \\
\midrule
\textbf{Model}
& \textbf{Cross-Signal Pred.}
& \textbf{Risk Assess.}
& \textbf{Signal Concord.}
& \textbf{Multimodal Phenotype}
& \textbf{Fitness Pred.}
& \textbf{Differential Diag.}
& \textbf{Prognostic Pred.}
& \textbf{Health Rec.} \\
\midrule
\rowcolor{highlight}
\multicolumn{9}{l}{\textit{Proprietary Models}} \\

GPT-4o
& 12.9$\pm$3.6
& 69.7$\pm$8.5
& 54.9$\pm$9.7
& 65.9$\pm$10.4
& 88.3$\pm$5.6
& 45.4$\pm$11.1
& 70.6$\pm$4.2
& 46.1$\pm$8.6 \\

GPT-5.4
& 25.8$\pm$4.7
& \underline{86.7$\pm$6.2}
& 77.5$\pm$8.0
& \underline{87.4$\pm$7.1}
& \underline{95.0$\pm$3.6}
& \underline{80.6$\pm$8.7}
& 68.6$\pm$4.3
& 53.1$\pm$8.6 \\

Claude Opus 4.6
& \textbf{40.5$\pm$5.3}
& 81.3$\pm$7.1
& \underline{78.5$\pm$7.9}
& \textbf{88.7$\pm$6.7}
& 78.2$\pm$7.4
& 63.7$\pm$10.7
& \underline{75.1$\pm$4.0}
& 51.6$\pm$8.7 \\

Claude Sonnet 4
& \underline{38.6$\pm$5.3}
& 85.8$\pm$6.4
& 69.6$\pm$8.9
& 77.3$\pm$9.2
& 84.1$\pm$6.5
& 59.8$\pm$11.0
& 73.1$\pm$4.1
& \underline{58.6$\pm$8.5} \\

Gemini 2.5 Pro
& 30.4$\pm$5.0
& 77.7$\pm$7.7
& 74.5$\pm$8.4
& 79.8$\pm$8.7
& 84.1$\pm$6.5
& 52.0$\pm$11.2
& 59.2$\pm$4.5
& 45.3$\pm$8.6 \\

Gemini 3.1 Pro
& 26.1$\pm$4.8
& \textbf{87.6$\pm$6.0}
& \textbf{85.3$\pm$6.7}
& 86.1$\pm$7.4
& \textbf{95.0$\pm$3.6}
& \textbf{80.6$\pm$8.7}
& \textbf{77.1$\pm$3.9}
& \textbf{61.7$\pm$8.4} \\

\hline
\rowcolor{highlight}
\multicolumn{9}{l}{\textit{Open-source Models}} \\

Llama-3.2-3B
& 12.3$\pm$3.5
& 33.9$\pm$8.8
& 27.4$\pm$8.6
& 26.5$\pm$9.7
& 44.5$\pm$8.9
& 32.4$\pm$10.4
& 42.3$\pm$4.6
& 40.6$\pm$8.5 \\

Llama-3.1-8B
& 8.3$\pm$2.9
& 43.7$\pm$9.2
& 39.2$\pm$9.5
& 39.2$\pm$10.8
& 56.3$\pm$8.9
& 37.6$\pm$10.8
& 49.2$\pm$4.6
& 41.4$\pm$8.5 \\

Llama-3.3-70B
& 8.6$\pm$3.0
& 53.6$\pm$9.2
& 35.3$\pm$9.3
& 54.4$\pm$11.0
& 80.7$\pm$7.0
& 42.8$\pm$11.1
& 47.0$\pm$4.6
& 49.2$\pm$8.7 \\

Gemma-3-4B
& 15.3$\pm$3.9
& 41.1$\pm$9.1
& 22.5$\pm$8.0
& 31.6$\pm$10.2
& 49.6$\pm$9.0
& 22.0$\pm$9.2
& 32.3$\pm$4.3
& 39.8$\pm$8.5 \\

Gemma-3-12B
& 14.7$\pm$3.8
& 58.0$\pm$9.1
& 35.3$\pm$9.3
& 54.4$\pm$11.0
& 70.6$\pm$8.2
& 22.0$\pm$9.2
& 47.0$\pm$4.6
& 42.2$\pm$8.6 \\

Gemma-3-27B
& 15.3$\pm$3.9
& 64.3$\pm$8.9
& 41.2$\pm$9.6
& 55.7$\pm$11.0
& 71.5$\pm$8.1
& 28.5$\pm$10.0
& 44.8$\pm$4.6
& 42.2$\pm$8.6 \\

Gemma-4-26B-A4B
& 22.7$\pm$4.5
& 73.2$\pm$8.2
& 54.9$\pm$9.7
& 77.3$\pm$9.2
& 91.7$\pm$4.8
& 68.9$\pm$10.3
& 71.7$\pm$4.2
& 56.3$\pm$8.6 \\

Mistral-Small-3.1
& 14.1$\pm$3.8
& 70.6$\pm$8.4
& 36.3$\pm$9.3
& 67.1$\pm$10.3
& 76.5$\pm$7.6
& 65.0$\pm$10.6
& 56.6$\pm$4.6
& 44.5$\pm$8.6 \\

\bottomrule
\end{tabular}
}

\end{table*}
\begin{table}[p]
\centering
\caption{\textbf{Comparison of chain-of-thought (CoT) and direct-answer (DA) prompting on \benchmark{}.}
Accuracy (\%) is reported for each model. Gap denotes CoT $-$ DA in percentage points, where positive values indicate an improvement from CoT prompting.}
\label{tab:cot-da}
\footnotesize
\begin{tabular}{lcc|c}
\toprule
\textbf{Model} & \textbf{CoT} & \textbf{DA} & \textbf{Gap} \\
\midrule
Gemini-3.1-Pro    & 72.9 & 71.3 & $+1.6$ \\
Claude-Opus-4.6   & 60.2 & 40.6 & $+19.6$ \\
Claude-Sonnet-4   & 53.2 & 48.3 & $+4.9$ \\
GPT-5.4           & 51.2 & 33.5 & $+17.7$ \\
Gemini-2.5-Pro    & 50.6 & 32.0 & $+18.6$ \\
Gemma-4-26B-A4B   & 42.5 & 35.7 & $+6.8$ \\
GPT-4o            & 34.7 & 23.9 & $+10.8$ \\
Llama-3.3-70B     & 29.2 & 27.2 & $+2.0$ \\
Mistral-Small-3.1 & 29.3 & 24.5 & $+4.8$ \\
Gemma-3-27B       & 26.2 & 23.8 & $+2.4$ \\
Gemma-3-12B       & 24.6 & 21.7 & $+2.9$ \\
Llama-3.1-8B      & 23.8 & 22.3 & $+1.5$ \\
Gemma-3-4B        & 19.8 & 20.6 & $-0.8$ \\
Llama-3.2-3B      & 19.6 & 17.9 & $+1.7$ \\
\bottomrule
\end{tabular}
\end{table}

\end{document}